\documentclass{article}
\usepackage[utf8]{inputenc}
\usepackage{amsmath,amssymb,amsthm,graphicx,rotating,color,enumerate,nicefrac}
\numberwithin{equation}{subsection}

\newenvironment{circlist}{
  \begin{enumerate}[$\circ$]
  \setlength{\itemsep}{0pt}
  \setlength{\parskip}{0pt}
  \setlength{\parsep}{0pt}
}{\end{enumerate}}

\DeclareMathSymbol{\leqslant}{\mathalpha}{AMSa}{"36}
\DeclareMathSymbol{\geqslant}{\mathalpha}{AMSa}{"3E}

\renewcommand{\ge}{\;\geqslant\;}

\def\O{\mathrm{O}}
\def\qq{\qquad}

\title{Spatially-sparse convolutional neural networks}
\author{
Benjamin Graham\\
{\small Dept of Statistics, University of Warwick, CV4 7AL, UK}\\
{\small \tt b.graham@warwick.ac.uk}\\
}

\begin{document}
\maketitle
\begin{abstract}
Convolutional neural networks (CNNs) perform well on problems such as handwriting recognition and image classification.
However, the performance of the networks is often limited by budget and time constraints, particularly when trying to train deep networks.

Motivated by the problem of online handwriting recognition, we developed a CNN for processing spatially-sparse inputs; a character drawn with a one-pixel wide pen on a high resolution grid looks like a sparse matrix. Taking advantage of the sparsity allowed us more efficiently to train and test large, deep CNNs. On the CASIA-OLHWDB1.1 dataset containing 3755 character classes we get a test error of 3.82\%.

Although pictures are not sparse, they can be thought of as sparse by adding padding. Applying a deep convolutional network using sparsity has resulted in a substantial reduction in test error on the CIFAR small picture datasets: 6.28\% on CIFAR-10 and 24.30\% for CIFAR-100.

\end{abstract}

\section{Introduction}
Convolutional neural networks typically consist of an input layer, a number of hidden layers, followed by a softmax classification layer. The input layer, and each of the hidden layers, is represented by a three-dimensional array with size, say, $M\times N\times N$. The second and third dimensions are spatial. The first dimension is simply a list of features available in each spatial location. For example, with RGB color images $N\times N$ is the image size and $M=3$ is the number of color channels.

The input array is processed using a mixture of convolution and pooling operations.
As you move forward through the network, $N$ decreases while $M$ is increased to compensate.
When the input array is spatially sparse, it makes sense to take advantage of the sparsity to speed up the computation. More importantly, knowing you can efficiently process sparse images gives you greater freedom when it comes to preparing the input data.

Consider the problem of online isolated character recognition; {\em online} means that the character is captured as a path using a touchscreen or electronic stylus, rather than being stored as a picture. Recognition of isolated characters can be used as a building block for reading cursive handwriting, and is a challenging problem in its own right for languages with large character sets.

Each handwritten character is represented as a sequence of strokes; each stroke is stored as a list of $x$- and $y$-coordinates. We can draw the characters as $N\times N$ binary images: zero for background, one for the pen color. The number of pixels is $N^2$, while the typical number of non-zero pixels is only $\O(N)$, so the first hidden layer can be calculated  much more quickly by taking advantage of sparsity.

Another advantage of sparsity is related to the issue of spatial padding for convolutional networks. Convolutional networks conventionally apply their convolutional filters in {\em valid} mode---they are only applied where they fit completely inside the input layer. This is generally suboptimal as makes it much harder to detect interesting features on the boundary of the input image. There are a number of ways of dealing with this.
\begin{circlist}
\item Padding the input image \cite{multicolumndeep} with zero pixels. This has a second advantage: training data augmentation can be carried out in the form of adding translations, rotations, or elastic distortions to the input images.
\item Adding small amounts of padding to each of the convolutional layers of the network; depending on the amount of padding added this may be equivalent to applying the convolutions in {\em full} mode. This has a similar effect to adding lots of padding to the input image, but it allows less flexibility when it comes to augmenting the training data.
\item Applying the convolutional network to a number of overlapping subsets of the image \cite{conf/nips/KrizhevskySH12}; this is useful if the input images are not square. This can be done relatively computationally efficiently as there is redundancy in the calculation of the lower level convolutional filters. However, the (often large) fully connected classification layers of the network must be evaluated several times.
\end{circlist}
Sparsity has the potential to combine the best features of the above.
The whole object can be evaluated in one go, with a substantial amount of padding added at no extra cost.

In Section \ref{deepcnet}-\ref{deepcnin} we describe a family of convolutional networks with many layers of max-pooling. In Section~\ref{sparsity}--\ref{nN} we describe how sparsity applies to character recognition and image recognition. In Section \ref{results} we give our results. In Section \ref{sec:conclusion} we discuss other possible uses of sparse CNNs.

\section{Deep convolutional networks and spatial sparsity}\label{DeepCNN}
Early convolutional networks tended to make use of pooling in a relatively restrained way, with two layers of pooling being quite typical. As computing power has become cheaper, the advantages of combining many layers of pooling with small convolutional filters to build much deeper networks have become apparent \cite{multicolumndeep,NetworkInNetwork}.

Applying pooling slowly, using many layers of $2\times 2$ pooling rather than a smaller number of $3\times 3$ or $4\times 4$ layers, may be particularly important for handwriting recognition.
The speed with which pooling is applied affects how a CNN will generalize from training data to test data. Slow max-pooling retains more spatial information; this is better for handwriting recognition, as handwriting is highly structured. Consider three small pictures:
\[
\overline{o} \qq \overline{o\vphantom{T}}
\qq \underline{o}
\]
Using fast max-pooling, all three pictures look the same: they all contain a circle and a line.
Using slower max-pooling, more spatial information is retained, so the first two pictures will be different from the third.

\begin{figure}
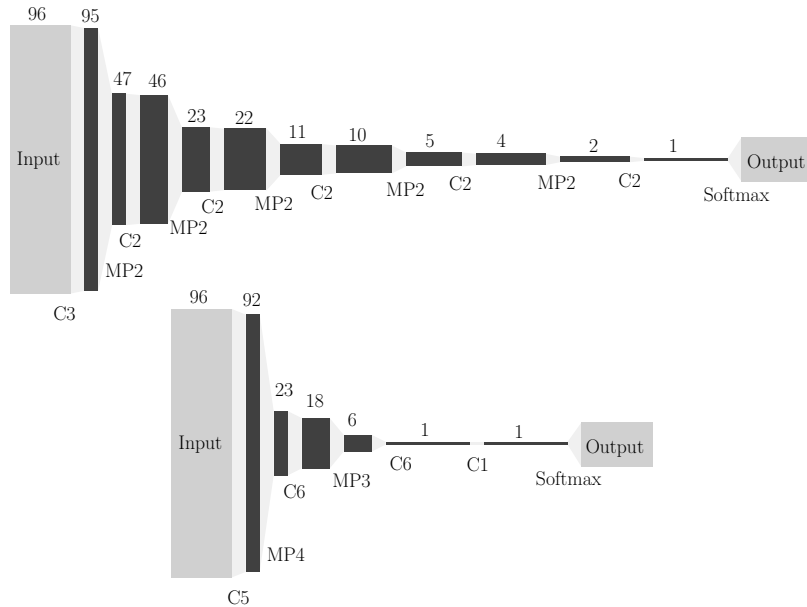

\begin{center}
\scalebox{0.3}{\input{a.pspdftex}}\\
\vspace{-5mm}
\scalebox{0.3}{\input{b.pspdftex}}
\end{center}
\caption{The hidden layer dimensions for $l=5$ DeepCNets, and LeNet-7. In both cases, the spatial sizes decrease from 96 down to 1. The DeepCNet applies max-pooling much more slowly than LeNet-7.\label{DeepCNetLeNet}}
\end{figure}

\begin{figure}
\resizebox{\columnwidth}{!}{
\mbox{
\includegraphics[width=2.88cm]{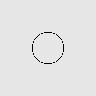}
\raisebox{0.03cm}{\includegraphics[width=2.82cm]{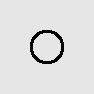}}
\raisebox{0.725cm}{\includegraphics[width=1.41cm]{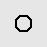}}
\raisebox{0.75cm}{\includegraphics[width=1.38cm]{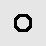}}
\raisebox{1.095cm}{\includegraphics[width=0.69cm]{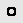}}
\raisebox{1.11cm}{\includegraphics[width=0.66cm]{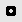}}
\raisebox{1.275cm}{\includegraphics[width=0.33cm]{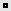}}
\raisebox{1.29cm}{\includegraphics[width=0.30cm]{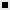}}
\raisebox{1.365cm}{\includegraphics[width=0.15cm]{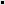}}
\raisebox{1.38cm}{\includegraphics[width=0.12cm]{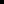}}
\raisebox{1.41cm}{\includegraphics[width=0.06cm]{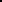}}
\raisebox{1.425cm}{\includegraphics[width=0.03cm]{circle11.png}}
}}
\caption{The active spatial locations (black) when a circle with diameter 32 is placed in the center of a DeepCNet$(5,\cdot)$ $96\times96$ input layer and fed forward. Sparsity is most important in the early stages where most of the spatial locations are in their ground state (gray). \label{circles}}
\end{figure}

For general input, slow pooling is relatively computationally expensive as the spatial size of the hidden layers reduces more slowly, see Figure~\ref{DeepCNetLeNet}. For sparse input, this is offset by the fact that sparsity is preserved in the early hidden layers, see Figure \ref{circles}. This is particularly important for handwriting recognition, which needs to operate on relatively low-power tablet computers.

\subsection{DeepCNet$(\ell,k)$}\label{deepcnet}
Inspired by \cite{multicolumndeep} we first consider a simple family of CNNs with alternating convolutional and max-pooling layers. We use two parameters $\ell$ and $k$ to characterize the structure:
there are $\ell+1$ layers of convolutional filters, separated by $\ell$ layers of $2\times2$ max-pooling.
The number of convolutional filters in the $n$-th convolutional layer is taken to be $nk$.
The spatial size of the filters is $3\times3$ in the first layer, and then $2\times2$ in the subsequent layers. Finally, at the top of the network is an output layer.

If the input layer has spatial size $N\times N$ with $N=3\times 2^\ell$, then the final convolution produces a fully connected hidden layer.
We call this network DeepCNet$(\ell,k)$.
For example, DeepCNet$(4,100)$ is the architecture from \cite{multicolumndeep} with input layer size $N=48=3\times 2^4$ and four layers of max-pooling:
\[
\text{input-100C3-MP2-200C2-MP2-300C2-MP2-400C2-MP2-500C2-output}
\]
For activation functions, we used the positive part function $f(x)=\max\{0,x\}$ for the hidden layers (rectified linear units) and softmax for the output layer.

The size of the hidden layers decays exponentially, so for general input the cost of evaluating the network is essentially just the cost of evaluating the first few layers: roughly speaking it is $\O(N^2k^2)$. Conversely, the number of trainable parameters of the model is
\[
3^2Mk+2^2k(k+1)+2^2(2k)(3k) +\dots+2^2(lk)((l+1)k)=\O(l^3k^2);
\]
most of the parameters are located in the top few layers.

When using dropout \cite{dropout} with a DeepCNets$(l,k)$, we need $l+2$ numbers to describe the amount of dropout applied to the input of the $l+1$ convolutional layers and the classification layer. The lower levels of a DeepCNet seem to be fairly immune to overfitting, but applying dropout in the upper levels improved test performance for larger values of $k$.


\subsection{DeepCNiN}\label{deepcnin}
Inspired by \cite{NetworkInNetwork}, we have tried modifying our DeepCNets by adding {\em network-in-network} layers. A NiN layer is a convolutional layer where the filters have spatial size just $1\times 1$.
They can be thought of as single layer networks that increase the learning power of a convolutional layer without changing the spatial structure. We placed NiN layers after each max-pooling layer and the final convolutional layer. With $k$ and $\ell$ as before, we call the resulting network DeepCNiN$(\ell,k)$. For example, DeepCNiN(4,100) is
\begin{align*}
\text{input}-&\text{100C3}-\text{MP2}-\textbf{100C1}-\dots\\
\dots-&\text{200C2}-\text{MP2}-\textbf{200C1}-\dots\\
\dots-&\text{300C2}-\text{MP2}-\textbf{300C1}-\dots\\
\dots-&\text{400C2}-\text{MP2}-\textbf{400C1}-\dots\\
\dots-&\text{500C2}-\textbf{500C1}-\text{output}
\end{align*}
As the spatial structure is unchanged, the cost of evaluating these networks is not much greater than for the corresponding DeepCNet. We took two steps to help the backpropagation algorithm operate effectively through such deep networks. First, we only applied dropout to the convolutional layers, not the NiN layers. Second, we used a form of {\em leaky} rectified linear units \cite{LEAKYRELU}, taking the activation function to be
\[
f(x)=
\begin{cases}
x,  &  x\ge 0,\\
x/3,  &  x<0.\\
\end{cases}
\]
Compared to \cite{LEAKYRELU}, we have used $x/3$ rather than $x/100$ for the $x<0$ case. This seems to speed up learning without harming the representational power of the network.

\subsection{Spatial sparsity for convolutional networks}\label{sparsity}
Imagine putting an all-zero array into the input layer of a CNN. As you evaluate the network in the forward direction, the translational invariance of the input is propagated to each of the hidden layers in turn. We can therefore think of each hidden variable as having a {\em ground state} corresponding to receiving no meaningful input; the ground state is generally non-zero because of bias terms. When the input array is sparse, you only have to calculate the values of the hidden variables where they differ from their ground state. Figure~\ref{circles} shows how the active spatial locations change through the layers.

Essentially, we want to memoize the convolutional and pooling operations.
Memoizing can be done using a hash table, but that would be inefficient here as for each operation there is only one input, corresponding to regions in the ground state, that we expect to see repeatedly.

Instead, to forward propagate the network we calculate two matrices for each layer of the network:
\begin{circlist}
\item A {\em feature matrix} which is a list of row vectors, one for the ground state, and one for each active spatial location in the layer;
the width of the matrix is the number of features per spatial location.
\item A {\em pointer matrix} with size equal to the spatial size of the convolutional layer. For each spatial location in the convolutional layer, we store the number of the corresponding row in the feature matrix.
\end{circlist}
This representation is very loosely biologically inspired. The human visual cortex separates into two streams of information: the dorsal ({\em where}) and ventral ({\em what}) streams.
Similar data structures can be used in reverse order for backpropagation.

For a regular convolutional network, the convolutional and pooling operations within a layer can be performed in parallel on a GPU (or even spread over multiple GPUs). Exactly the same holds here; there is simply less work to do as we know the inactive output spatial locations are all the same.

\subsection{Online Character Recognition}\label{onlinecr}
Online character recognition is the task of reading characters represented as a collection of pen-stroke paths. Two rather different techniques work particularly well for online Chinese character recognition.
\begin{circlist}
\item Render the pen strokes at a relatively high resolution, say $n\times n$ with $n=40$, and then use a CNN as a classifier \cite{multicolumndeep}.
\item Draw the character in a much lower resolution grid, say $n\times n$ with $n=8$. In each square of the grid calculate an 8-dimensional {\em histogram} measuring the amount of movement in each of the 8 compass directions. The resulting $8n^2=512$ dimensional vectors are suitable input for general purpose statistical classifiers \cite{bb104561}.
\end{circlist}
The first representation records more accurately {\em where} the pen went, while the second is better at recording the {\em direction} the pen was taking.
Using sparsity, we can try to get the best of both worlds. Combining the two representations gives an array of size $(1+8)\times n\times n$. Setting $n=64$ gives a sparse representation of the character suitable for feeding into a CNN.
This preserves sparsity as the histogram is all zero at sites the pen does not touch. Increasing the number of input features per spatial location only increases the cost of evaluating the first hidden layer, so for sparse input it tends to have a negligible impact on performance.

The idea of supplementing pictures of online pen strokes with extra features has been used before, for example in the context of cursive English handwriting recognition \cite{lecun-bengio-94}. The key difference to previous work is the use of sparsity to allow a substantial increase in the spatial resolution, allowing us to obtain good results for challenging datasets such as CASIA-OLHWDB1.1.

\subsection{Offline/Image recognition}\label{offlinecr}
In contrast to online character recognition, for offline character recognition you simply have a picture of each character---for example, the MNIST digit dataset \cite{mnistlecun}. The $28\times28$ pictures have on average 150 non-zero pixels.

The input layer for a DeepCNet$(5,\cdot)$ network is $96\times 96$. Placing the MNIST digits in the middle of the input layer produces a sparse dataset as $150$ is much smaller than $96^2$. The background of each MNIST picture is zero, so extending the images simply increases the size of the background.

To a lesser extent, we can also think of the $32\times 32$ images in the CIFAR-10 and CIFAR-100 datasets as being sparse, again by placing them into a $96\times 96$ grid. For the CIFAR datasets, we scale the RGB features to the interval $[-1,1]$. Padding the pictures with zeros corresponds to framing the images with gray pixels.

\subsection{Object scale versus input scale}\label{nN}
Let $n$ denote the size of the sparse objects to be recognized.
For offline/image recognition, this is the width of the images.
For online character recognition, we are free to choose a scale $n$ on which to draw the characters.

\begin{figure}
\begin{center}
\resizebox{0.48\columnwidth}{!}{\includegraphics[width=6.4cm,trim=2.5cm 8cm 2.5cm 8cm,clip]{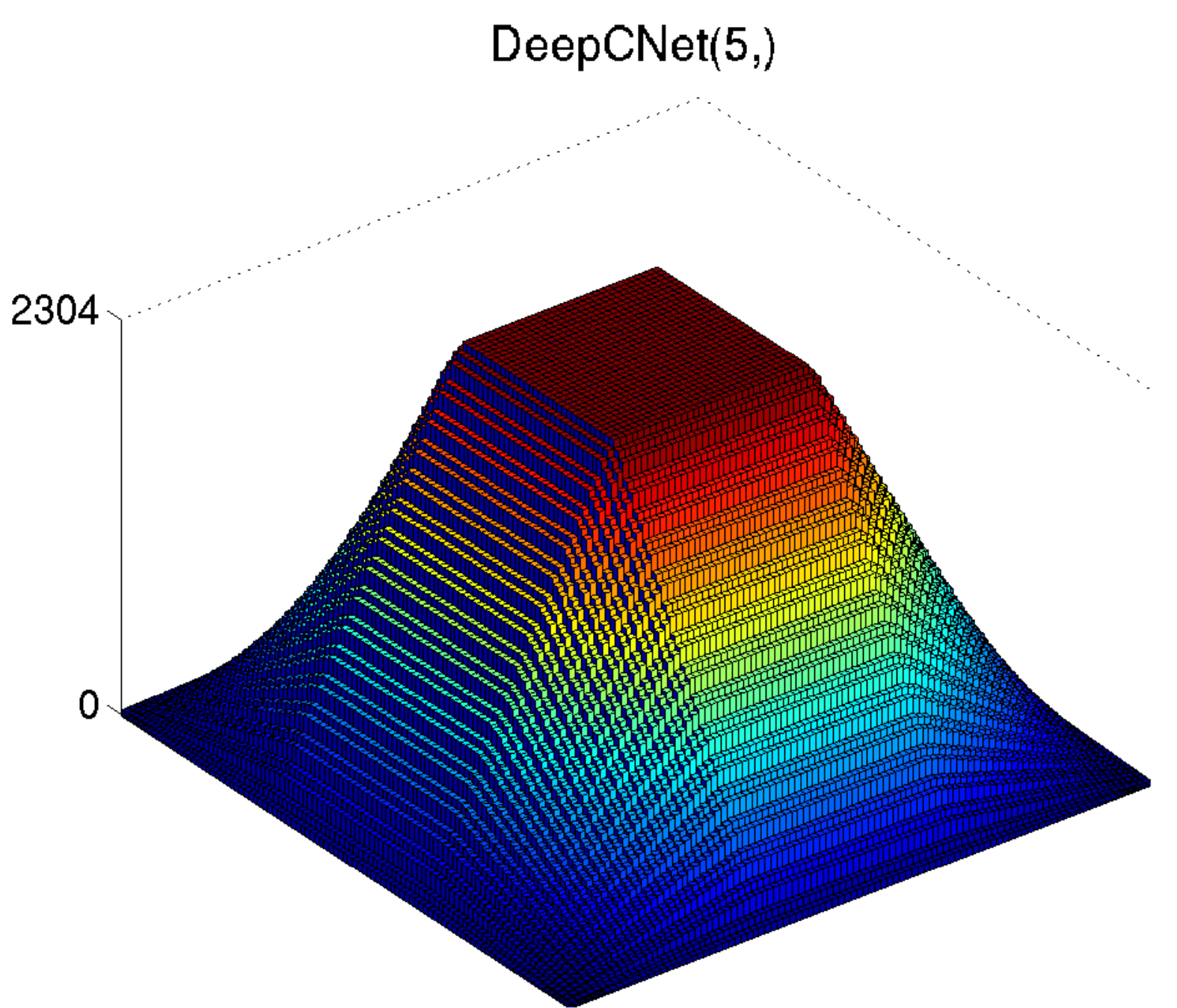}}\resizebox{0.48\columnwidth}{!}{\includegraphics[width=6.4cm,trim=2.5cm 8cm 2.5cm 8cm,clip]{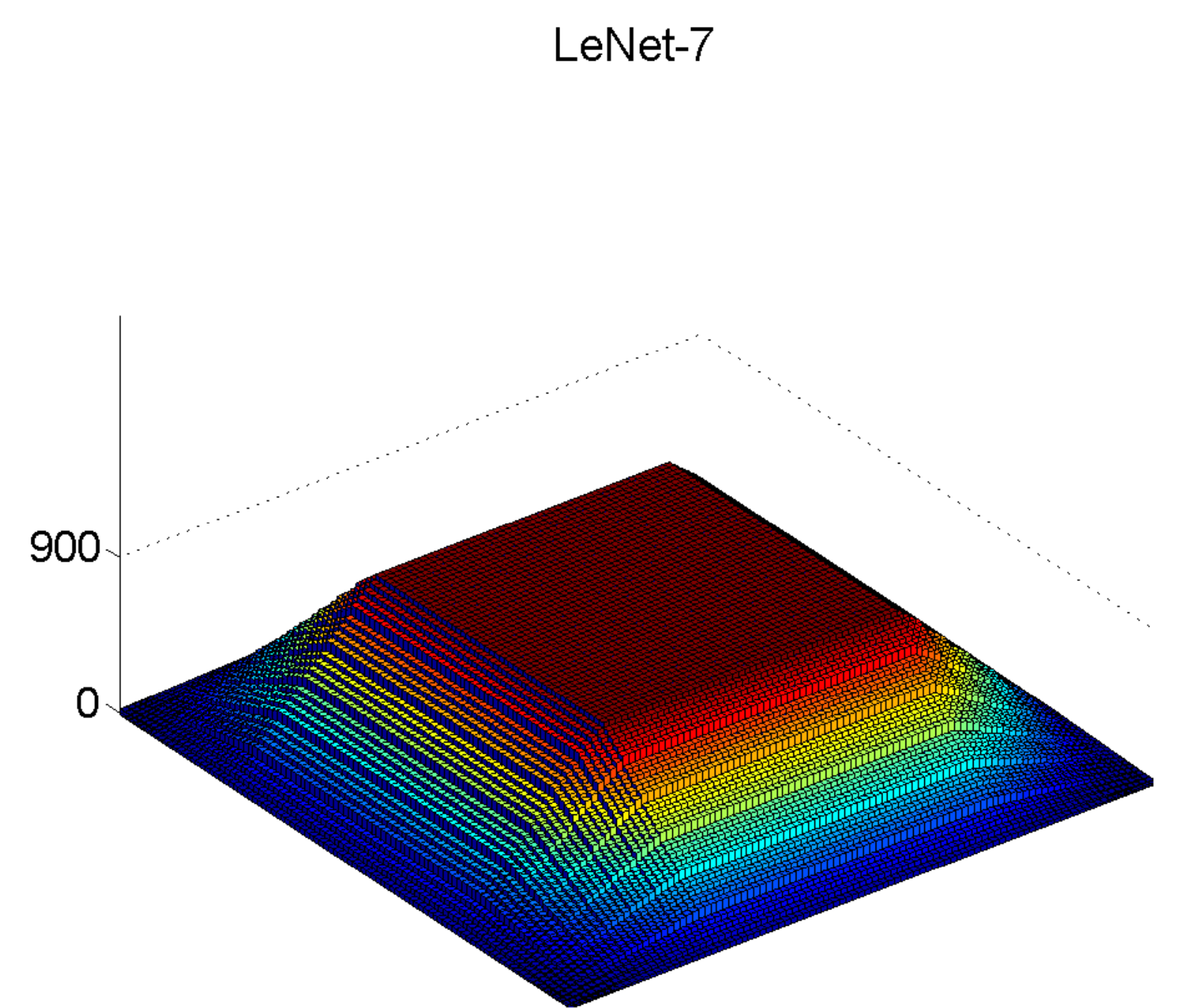}}
\caption{Top: A comparison of the number of possible paths from the input layer ($96\times 96$) to the fully connected layer for $\ell=5$ DeepCNets and LeNet-7 \cite{lecun-04}. The DeepCNet's larger number of hidden layers translates into a larger maximum number of paths, but with a narrower plateau.
\label{counts}}
\end{center}
\end{figure}

Given $n$, we must choose the $\ell$-parameter such that the characters fit comfortably into the $N\times N$ input layer, where $N=3\times 2^\ell$. DeepCNets work well with $n\approx N/3$. There are a number of ways to account for this:
\begin{circlist}
\item To process the $n\times n$ sized input down to a zero-dimensional quantity, the number $\ell=\log_2(N/3)$ of levels of $2\times2$ max-pooling should be approximately $\log_2 n$.
\item Counting the number of paths through the CNN from the input to output layers reveals a plateau; see Figure~\ref{counts}. Each corner of the input layer has only one route to the output layer; in contrast, the central $(N/3)\times(N/3)$ points in the input layer each have $3^2\times 2^{2(\ell-1)}$ such paths.
\item If the input-layer is substantially larger than the object to be recognized, then you can apply various transformations (translations, rotations, etc) to the training images without having to worry about truncating the input.
\item When the input is sparse, translating the input by $2^k$ corresponds to translating the output of the $k$-th level of $2\times2$ max-pooling by exactly one. Adding moderately large random shifts to the training data forces the network to learn how the alignment of input with the max-pooling layers affects the flow of information through the network.
\end{circlist}

\section{Results}\label{results}
The results are all test errors for single networks.
We consider two ways of augmenting the training data:
\begin{circlist}
\item By translation, adding random shifts in the $x$ and $y$ directions.
\item By affine transformations, using a randomized mix of translations, rotations, stretching and shearing operations.
\end{circlist}
In Section \ref{varyl} we see that it is crucial to move the training data around inside the much larger input field. We guess that this is an unavoidable consequence of the slow max-pooling.

It is therefore difficult to compare meaningfully DeepCNet test results with the results of other papers, as they often deliberately limit the use of data augmentation to avoid the temptation of learning from the test data. As a compromise, we restrict ourselves to only extending the MNIST dataset by translations. For other datasets, we use affine transformations, but we do not use elastic deformations or cropped versions of the training images.

Some language specific handwriting recognition systems preprocess the characters to truncate outlying strokes, or apply local scalings to make the characters more uniform. The aim is to remove variation between writing styles, and to make the most of an input layer that is limited in size. However, there is a danger that useful information is being discarded. Our method seems able to handle a range of datasets without preprocessing. This seems to be a benefit of the large input grid size and the use of slow max-pooling.

\subsection{Small online datasets: 10, 26 and 183 characters}\label{small-datasets}
\begin{figure}
\resizebox{\columnwidth}{!}{\includegraphics[width=125mm,trim=0cm 0cm 21cm 0cm,clip]{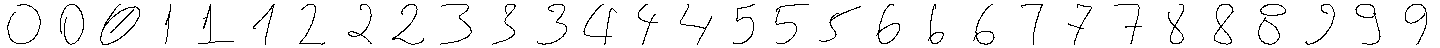}}\\

\resizebox{\columnwidth}{!}{\includegraphics[width=125mm,trim=0cm 0cm 21cm 0cm,clip]{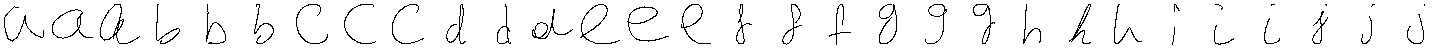}}\\

\resizebox{\columnwidth}{!}{\includegraphics[width=125mm,trim=0cm 0cm 21cm 0cm,clip]{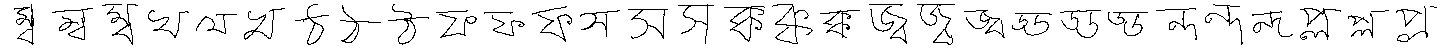}}\\

\resizebox{\columnwidth}{!}{\includegraphics[width=125mm,trim=0cm 0cm 42cm 0cm,clip]{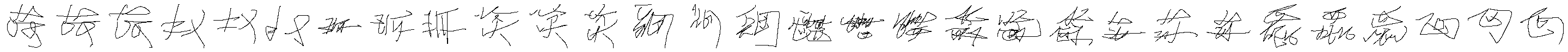}}
\caption{Samples from the Pendigits, UJIpenchars and Assamese datasets (resolution $40\times 40$) and from the CASIA dataset ($80\times80$).\label{datasets}}
\end{figure}

We will first look at three relatively small datasets \cite{UCIrep} to study the effect of varying the network parameters. See Figure~\ref{datasets}.
\begin{circlist}
\item The Pendigits dataset contains handwritten digits 0-9. It contain about 750 training characters for each of the ten classes.
\item The UJIpenchars database includes the 26 lowercase letters. The training set contains 80 characters for each of the 26 classes.
\item The Online Handwritten Assamese Characters Dataset contains 45 samples of 183 Indo-Aryan characters. We used the first 36 handwriting samples as the training set, and the remaining 9 samples for a test set.
\end{circlist}
As described in Section \ref{nN}, when using a DeepCNet$(\ell,k)$ we draw the characters with size $2^\ell\times2^\ell$ in the center of the input layer of size $(3\times 2^\ell)\times(3\times 2^\ell)$.
As described in Section \ref{onlinecr}, the number of features $M$ per spatial location is either 1 (simple pictures) or 9 (pictures and 8-directional histogram features).

\subsubsection{Varying $\ell$}\label{varyl}
We first explore the impact of increasing the number $\ell$ of layers of max-pooling; equivalently we are increasing the scale $2^\ell$ at which the characters are drawn. For simplicity we only look here at the Assamese dataset. In terms of computational cost, increasing $\ell$ by one approximately doubles the number of active input spatial locations. When $l=6$, characters are drawn in a $64\times 64$ square, and about 6\% of the spatial sites are active.

With only 36 handwriting samples for training, we should expect the dataset to be quite challenging, given the large alphabet size, and the relatively intricate characters. We see in Table~\ref{assameseResults} that adding translations to the training data is clearly a crucial part of training DeepCNets. Adding more general affine transforms also helps.

\begin{table}[t]
\begin{center}
\begin{tabular}{|c|c|c|c|}
\hline
DeepCNet$(\ell,30)$ & None       & Translations & Affine\\
\hline
$\ell=3$            & 51.4\%     & 38.3\% & 35.2\% \\
$\ell=4$            & 26.8\%     & 9.29\% & 7.47\% \\
$\ell=5$            & 18.2\%     & 5.28\% & 2.73\% \\
$\ell=6$            & 14.1\%     & 4.61\% & 1.76\% \\
$\ell=7$            & \bf{13.8\%}& \bf{4.07\%} &\bf{1.70\%}\\
\hline
\end{tabular}
\end{center}
\caption{Test errors for the Assamese dataset for different values of $\ell$ and different amounts of data augmentation ($M=1$).\label{assameseResults}}
\end{table}

\subsubsection{Adding 8-direction histogram features}
Here we look at the effect of increasing the feature set size. To make the comparisons interesting, we deliberately restrict the DeepCNet $k$ and $\ell$ parameters. Increasing $k$ or $\ell$ is computationally expensive. In contrast, increasing $M$ only increases the cost of evaluating the first layer of the CNN; in general for sparse CNNs the first layer represents only a small fraction of the total cost of evaluating the network. Thus increasing $M$ is cheap, and we will see that it tends to improve test performance.

We trained DeepCNet$(4,20)$ networks, extending the training set by either just translations or affine transforms.
In Table~\ref{M1_9} we see that adding extra features can improve generalization.

\begin{table}[t]
\begin{center}
\noindent\begin{tabular}{|cc|c|c|}
\hline
Dataset     & augmented by & Pictures  & Pictures \& Histograms\\
            &              & $M=1$     & $M=9$\\
\hline 
Pendigits   & Translations & 1.5\%     & \bf{0.74\%}\\
Pendigits   & Affine       & 0.91\%    & \bf{0.69\%}\\
UJIpenchars & Translations & 8.6\%     & \bf{ 7.6\%} \\
UJIpenchars & Affine       & 6.8\%     & \bf{ 6.5\%} \\
Assamese    & Translations & 15.2\%    & \bf{ 7.7\%} \\
Assamese    & Affine       & 13.5\%    & \bf{ 4.9\%} \\
\hline
\end{tabular}
\end{center}
\caption{Test errors with and without adding 8-directional histogram features to the input representation.\label{M1_9}}
\end{table}

\subsubsection{CASIA: 3755 character classes}
The CASIA-OLHWDB1.1\cite{CASIA} database contains online handwriting samples of the 3755 GBK level-1 Chinese characters. There are approximately 240 training characters, and 60 test characters, per class. A test error of 5.61\% is achieved in \cite{multicolumndeep} using a DeepCNet(4,100) applied to $40\times 40$ pictures drawn from the pen data.

We trained a smaller network without dropout, and a larger network with dropout per level of
 $0,0,0,0.1,0.2,0.3,0.4,0.5$. In both cases we extended the dataset with affine transformations. In Table~\ref{casiaTable} we see that sparsity allows us to get pretty good performance at relatively low computational cost, and good performance with a larger network.

\begin{table}[t]
\begin{center}
\begin{tabular}{|c|c|c|}
\hline
Network  &Pictures    & Pictures \& Histograms \\
         &$M=1$ & $M=9$\\
\hline
DeepCNet(6,30)  & 9.69\% & \bf{6.63}\%\\
\hline
DeepCNet(6,100)  & 5.12\% & \bf{3.82}\%\\
\hline
\end{tabular}
\end{center}
\caption{Test errors for CASIA-OLHWDB1.1.\label{casiaTable}}
\end{table}

\subsubsection{ICDAR2013 Chinese Handwriting Recognition Competition Task 3}
We entered a version of our program into the ICDAR2013 competition for task 3, trained on the CASIA OLHWDB1.0-1.2 datasets. Our entry won with a test error of 2.61\%, compared to 3.13\% for the second best entry. Human performance on the test dataset was reported to be 4.81\%. Our ten-best-guesses included the correct character 99.88\% of the time.

\subsection{MNIST digits}\label{mnist}
Let us move now from online to offline handwriting recognition. We extended the MNIST training set by translations only, with shifts of up to $\pm2$ pixels. We first trained a very small-but-deep network, DeepCNet$(5,10)$.  This gave a test error of 0.58\%. Using a NVIDIA GeForce GTX 680 graphics card, we can classify 3000 characters per second.

Training a DeepCNet$(5,60)$ with dropout per level of $0,0,0,0.5,0.5,0.5,0.5$ resulted in a test error of 0.31\%.
Better results on the MNIST dataset have been obtained, but only by training committees of neural networks, and by using elastic distortions to augment the dataset.

\subsection{CIFAR-10 and CIFAR-100 small pictures}
These two dataset each contain 50,000 $32\times32$ color training images, split between 10 and 100 categories, respectively. For each of the datasets, we extended the training set by affine transformations, and trained a DeepCNet(5,300) network with dropout per convolutional level of $0,0,0.1,0.2,0.3,0.4,0.5$.
The resulting test errors were 8.37\% and 29.81\%. For comparison \cite{NetworkInNetwork} reports errors of 8.81\% (with data augmentation) and 35.68\% (without data augmentation).

Adding Network in Network layers to give a DeepCNiN(5,300) network produced substantially lower test errors:  6.28\% and 24.30\%, respectively.

\section{Conclusion}\label{sec:conclusion}
We have implemented a spatially-sparse convolutional neural network, and shown that it can be applied to the problems of handwriting and image recognition. Sparsity has allowed us to use CNN architectures that we would otherwise have rejected as being too slow.

For online character recognition, we have extended the dense 8-directional histogram representation to produce a sparse representation. There are other ways of creating sparse representation, for example by using path curvature as a feature. Trying other sparse representation could help improve performance.

Very recently, convolutional networks somewhat similar to our DeepCNiN, but trained with multiple GPUs or networks of computers, have been shown to perform well on the LSVRC2012 image dataset \cite{VGG2014, GoogLeNet}. For example, a network from \cite{VGG2014} uses layers of type C3-C3-MP2 as opposed to our C2-MP2-C1 DeepCNiN layers. Running larger, sparse CNNs could potentially improve accuracy or efficiency by making it practical to train on uncropped images, and with greater flexibility for training set augmentation.

Sparsity could also be used in combination with a segmentation algorithm. Segmentation algorithms divide an image into irregularly shaped foreground and background elements. A sparse CNN could be applied to classify efficiently the different segments. Also, by mixing and matching neighboring segments, a sparse CNN could be used to check if two segments are really both parts of a larger object.

Sparsity could also be used to implement higher dimensional convolutional networks, where the convolutional filters range over three or more dimensions \cite{3dConvolutions}. Higher dimensional convolutions can be used to analyze static objects in 3D, or objects moving in 2+1 or 3+1 dimensional space-time. Just as lines form sparse sets when embedded in two or higher dimensional spaces, surfaces form sparse sets when embedded in three or higher dimensional spaces. Possible applications include analyzing 3D representations of human faces (2D surfaces in 3D space) or analyzing the paths of airplanes (1D lines in 4D space-time).

\section{Acknowledgement}
Many thanks to Fei Yin, Qiu-Feng Wang and Cheng-Lin Liu for their work organizing the ICDAR2013 competition.


\begin{thebibliography}{10}

\bibitem{multicolumndeep}
D.~Ciresan, U.~Meier, and J.~Schmidhuber.
\newblock Multi-column deep neural networks for image classification.
\newblock In {\em Computer Vision and Pattern Recognition (CVPR), 2012 IEEE
  Conference on}, pages 3642--3649, 2012.

\bibitem{conf/nips/KrizhevskySH12}
Alex Krizhevsky, Ilya Sutskever, and Geoffrey~E. Hinton.
\newblock Imagenet classification with deep convolutional neural networks.
\newblock In Peter~L. Bartlett, Fernando C.~N. Pereira, Christopher J.~C.
  Burges, Léon Bottou, and Kilian~Q. Weinberger, editors, {\em NIPS}, pages
  1106--1114, 2012.

\bibitem{NetworkInNetwork}
Min Lin, Qiang Chen, and Shuicheng Yan.
\newblock Network in network.
\newblock {\em CoRR}, abs/1312.4400, 2013.

\bibitem{dropout}
Geoffrey~E. Hinton, Nitish Srivastava, Alex Krizhevsky, Ilya Sutskever, and
  Ruslan Salakhutdinov.
\newblock Improving neural networks by preventing co-adaptation of feature
  detectors.
\newblock {\em CoRR}, abs/1207.0580, 2012.

\bibitem{LEAKYRELU}
Andrew~L. Maas, Awni~Y. Hannun, and Andrew~Y. Ng.
\newblock Rectifier nonlinearities improve neural network acoustic models.
\newblock volume ICML WDLASL 2013.

\bibitem{bb104561}
Z.~L. Bai and Q.~A. Huo.
\newblock A study on the use of 8-directional features for online handwritten
  {C}hinese character recognition.
\newblock In {\em ICDAR}, pages I: 262--266, 2005.

\bibitem{lecun-bengio-94}
Y.~Le{C}un and Y.~Bengio.
\newblock word-level training of a handwritten word recognizer based on
  convolutional neural networks.
\newblock In IAPR, editor, {\em Proc. of the International Conference on
  Pattern Recognition}, volume~II, pages 88--92, Jerusalem, October 1994. IEEE.

\bibitem{mnistlecun}
Yann Le{C}un and Corinna Cortes.
\newblock {The MNIST database of handwritten digits}.
\newblock http://yann.lecun.com/exdb/mnist/.

\bibitem{lecun-04}
Yann Le{C}un, Fu-Jie Huang, and Leon Bottou.
\newblock Learning methods for generic object recognition with invariance to
  pose and lighting.
\newblock In {\em Proceedings of CVPR'04}. IEEE Press, 2004.

\bibitem{UCIrep}
K.~Bache and M.~Lichman.
\newblock {UCI} machine learning repository, 2013.

\bibitem{CASIA}
C.-L. Liu, F.~Yin, D.-H. Wang, and Q.-F. Wang.
\newblock {CASIA} online and offline {C}hinese handwriting databases.
\newblock In {\em Proc. 11th International Conference on Document Analysis and
  Recognition (ICDAR), Beijing, China}, pages 37--41, 2011.

\bibitem{VGG2014}
Karen Simonyan and Andrew Zisserman.
\newblock Very deep convolutional networks for large-scale image recognition.
\newblock {\em arxiv.org/abs/1409.1556}.

\bibitem{GoogLeNet}
Christian Szegedy, Wei Liu, Yangqing Jia, Pierre Sermanet, Scott Reed, Dragomir
  Anguelov, Dumitru Erhan, Vincent Vanhoucke, and Andrew Rabinovich.
\newblock Going deeper with convolutions.
\newblock {\em http://arxiv.org/abs/1409.4842}.

\bibitem{3dConvolutions}
Shuiwang Ji, Wei Xu, Ming Yang, and Kai Yu.
\newblock 3d convolutional neural networks for human action recognition.
\newblock {\em IEEE Trans. Pattern Anal. Mach. Intell.}, 35(1):221--231,
  January 2013.

\end{thebibliography}
\end{document}